\documentclass[letterpaper, 10 pt, conference]{ieeeconf}  

\IEEEoverridecommandlockouts                              
\usepackage{amsfonts}
\usepackage{amsmath}
\usepackage{booktabs}
\usepackage{graphicx}
\usepackage{hyperref}
\usepackage{multirow}
\usepackage{tabularx}
\usepackage{adjustbox}

\usepackage{subcaption}
\usepackage{hyperref}
\usepackage{listings}
\usepackage{xcolor} 

\usepackage[style=ieee]{biblatex}
\usepackage{etoolbox}
\AtBeginBibliography{\small}

\title{\LARGE \bf 
    OSMa-Bench: Evaluating Open Semantic Mapping Under Varying Lighting Conditions
}

\author{
    Maxim Popov$^{1*}$, Regina Kurkova$^{1}$, Mikhail Iumanov$^{1}$, Jaafar Mahmoud$^{1}$, and Sergey Kolyubin$^{1}$
    \thanks{$^{1}$ Biomechatronics and Energy-Efficient Robotics (BE2R) Lab, ITMO University, Saint Petersburg, Russia}
    \thanks{$^{*}$ Corresponding author: mfpopov@itmo.ru}%
    \thanks{This work was written with the financial support from project no. FSER-2025-0002.}
}

\begin{document}
\maketitle
\thispagestyle{empty}
\pagestyle{empty}


\begin{abstract}
    Open Semantic Mapping (OSM) is a key technology in robotic perception, combining semantic segmentation and SLAM techniques. This paper introduces a dynamically configurable and highly automated LLM/LVLM-powered pipeline for evaluating OSM solutions called \textit{OSMa-Bench} (Open Semantic Mapping Benchmark). The study focuses on evaluating state-of-the-art semantic mapping algorithms under varying indoor lighting conditions, a critical challenge in indoor environments. We introduce a novel dataset with simulated RGB-D sequences and ground truth 3D reconstructions, facilitating the rigorous analysis of mapping performance across different lighting conditions. Through experiments on leading models such as ConceptGraphs~\cite{gu2024conceptgraphs}, BBQ~\cite{linok2024beyond}, and OpenScene~\cite{peng2023openscene}, we evaluate the semantic fidelity of object recognition and segmentation. Additionally, we introduce a scene graph evaluation method to analyze the ability of models to interpret semantic structure. The results provide insights into the robustness of these models, forming future research directions for developing resilient and adaptable robotic systems. Our code is available at \url{https://be2rlab.github.io/OSMa-Bench/}.
\end{abstract}

\section{Introduction}

Open-vocabulary semantic mapping (OSM) is a rapidly advancing field in robotics, computer vision, and natural language processing. OSM enables robots to construct rich, semantically meaningful representations of their environment, empowering them to navigate and interact more efficiently. Recent state-of-the-art approaches have made impressive strides in leveraging vision-language models (VLMs) and novel 3D scene reconstruction techniques to generate highly expressive and interpretable spatial representations, making robots for general purposes one step closer \cite{gu2024conceptgraphs, linok2024beyond, peng2023openscene, jatavallabhula2023conceptfusion, werby2024hierarchical, yamazaki2024open, takmaz2023openmask3d, nguyen2024open3dis, ding2023pla, ha2022semantic, huang2024openins3d, chang2023context, martins2024ovo}. However, several challenges remain to be addressed to fully utilize OSM in real-world robotics applications. These include maintaining consistency in local updates, improving the clarity of semantic boundaries, adapting models to specific domains, increasing computational efficiency, and resolving ambiguities in natural language. Critically, many VLMs are trained on large-scale web datasets of natural images, yet robotic environments often exhibit significantly different visual characteristics. Indoor settings can experience dramatic variations in illumination throughout the day and from artificial lighting. Domain-specific visual artifacts introduce additional complications. For OSM systems to be practical, they must be able to handle these diverse and dynamic conditions robustly.

\begin{figure}[t]
    \centering
    \includegraphics[width=\columnwidth]{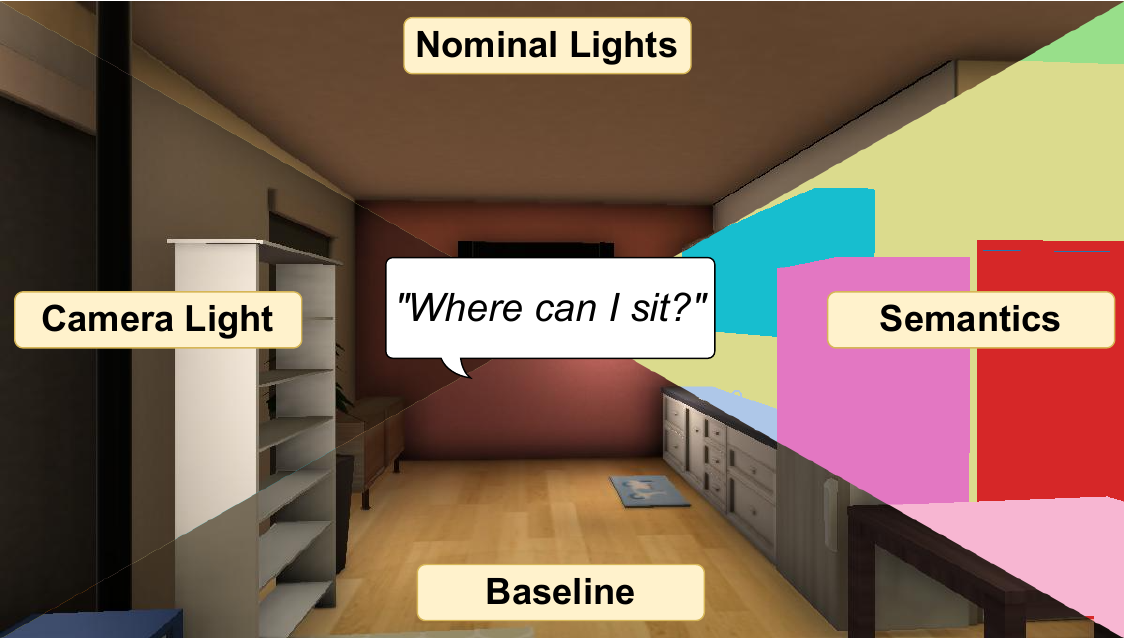}
    \caption{Our work evaluates open semantic mapping quality, providing an automated LLM/LVLM-based alternative to human assessment. We generate test sequences with different lighting conditions in a simulated indoor environment. Extra modifier, such as variations in robot nominal velocity, is applied as well.}
    \label{fig:teaser}
\end{figure}

\section{Related Work}

Recent progress in open-vocabulary semantic mapping leverages VLMs and multimodal fusion to enable robots to interpret scenes via natural language. Existing approaches fall into three categories: 3D scene understanding, language-guided segmentation, and robustness-oriented mapping.
\paragraph{Open-Vocabulary 3D Scene Understanding}
Most methods adopt 2D-3D feature fusion to project CLIP~\cite{radford2021learning} features in 3D. ConceptFusion~\cite{jatavallabhula2023conceptfusion} aggregates image features into point clouds to enable semantic queries, while ConceptGraphs~\cite{gu2024conceptgraphs} and HOV-SG~\cite{werby2024hierarchical} extend this by constructing hierarchical scene graphs that capture object relations. OpenFusion~\cite{yamazaki2024open} improves runtime efficiency, but emphasizes FPS over robustness. These methods are commonly benchmarked on ScanNet~\cite{dai2017scannet} or Replica~\cite{straub2019replica} datasets with controlled lighting, limiting generalization to dynamic environments.

\paragraph{Language-Guided 3D Instance Segmentation}
Some approaches focus on grounding language in 3D geometry. OpenScene~\cite{peng2023openscene} co-embeds point features with CLIP for zero-shot segmentation, while OpenMask3D~\cite{takmaz2023openmask3d} and Open3DIS~\cite{nguyen2024open3dis} fuse class-agnostic 3D masks with 2D classifiers. Despite strong performance, these models assume stable sensor and lighting conditions, which are rarely met in real-world applications.
\paragraph{Robustness-Oriented Semantic Mapping}
Many works in the literature target environmental variability. BBQ~\cite{linok2024beyond} uses LLMs with 3D scene graphs to resolve ambiguous queries, though with high compute costs. PLA~\cite{ding2023pla} introduces planar constraints to enforce geometric consistency, and Semantic Abstraction~\cite{ha2022semantic} applies LLM-driven filtering to reduce noise. However, evaluations still lack low-light or dynamic condition testing.

The evaluation methodologies of the current benchmarks focus narrowly on segmentation accuracy (mIoU, mAP50) and geometric fidelity, neglecting operational factors critical for robotics. For example, OpenIns3D~\cite{huang2024openins3d} evaluates instance segmentation across five datasets using AP25/AP50 metrics but does not quantify performance degradation under sensor noise or motion blur. Existing validation of graph scenes representations rarely goes beyond grounding: OVSG~\cite{chang2023context} introduces a grounding success rate for spatial queries but assumes static illumination and lacks standardized robustness protocols for lighting variation and sensor dynamics; in practice, evaluations rely on manually curated datasets or human assessment. Outside mapping, structured 3D‑QA benchmarks such as GQA~\cite{hudson2019gqa} and Space3D‑Bench~\cite{szymanska2024space3d} are commonly used, but still not in an automatic way.

We address this gap by evaluating four critical robotic capabilities: semantic segmentation, object existence, detection completeness, and spatial reasoning - under conditions that simulate real-world uncertainty.

\begin{figure*}[htbp]
    \centering
    \includegraphics[width=\textwidth]{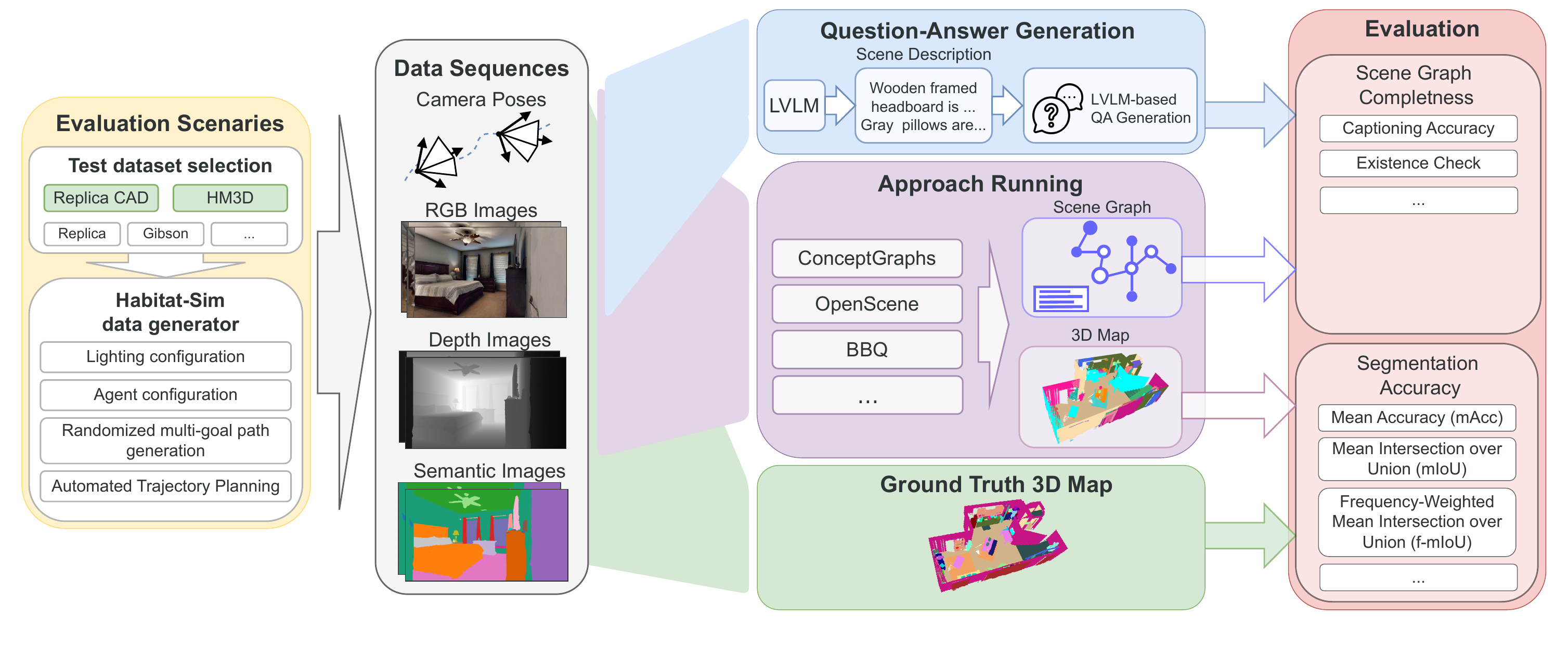}
    \caption{\textit{Evaluation Pipeline Diagram}.  We use Habitat Sim to establish test configurations that include various environmental factors like lighting and agent trajectories, generating a diverse dataset. This allows for tailored scenes, with randomly initialized agent trajectories simulating realistic interactions for effective model evaluation.}
    \label{fig:full_pipeline}
\end{figure*}
\section{Main Contribution}

\begin{figure}[ht]
    \centering
    \includegraphics[width=\columnwidth]{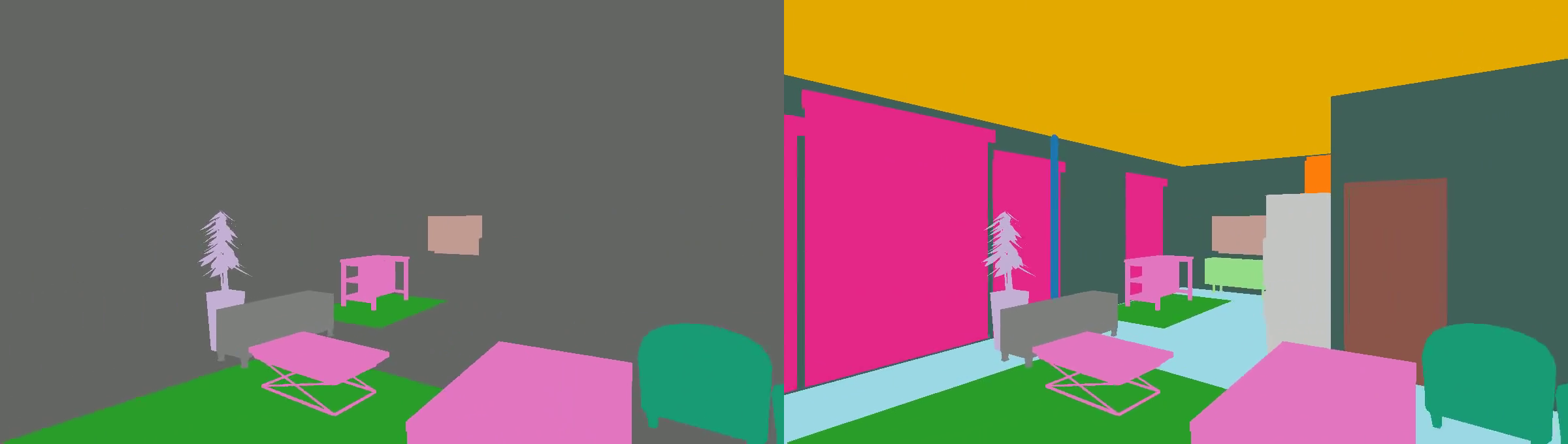}
    \caption{\textit{Replica CAD Dataset with Augmented Semantics}. We expanded the semantic description of the replica CAD dataset. This made it possible to take into account during testing both classes describing parts of the apartment (for example, ``\texttt{wall}'', ``\texttt{floor}'', ``\texttt{stairs}'') and classes describing furniture and household utensils.}
    \label{fig:augmented_replica_cad_semantics}
\end{figure}

\textit{OSMa-Bench (Open Semantic Mapping Benchmark)} introduces a novel framework for evaluating robustness in open semantic mapping, addressing gaps in an indoor environment simulation and multimodal reasoning assessment. Our key contributions are:
\begin{enumerate}
\item \textbf{OSMa-Bench Datasets}: ReplicaCAD~\cite{szot2021habitat} and HM3D~\cite{ramakrishnan2021habitat} extended with lighting modes and doubled/nominal robot speed, plus per-instance masks (Fig.~\ref{fig:augmented_replica_cad_semantics}).
\item \textbf{LLM-based QA Pipeline}: Gemini~\cite{team2024gemini} generates and grades graph-aware questions, yielding fully automated semantic evaluation.
\item \textbf{Cross-Method Robustness Analysis}: We compare OpenScene, ConceptGraphs, and BBQ \cite{peng2023openscene, gu2024conceptgraphs, linok2024beyond}, showing the effect of case conditions on their accuracy and reasoning capabilities.
\end{enumerate}

\textit{OSMa-Bench} unifies dynamic simulation, automated semantic validation, and reproducible evaluation, providing actionable insights for developing robust OSM systems.

Unlike static benchmarks, OSMa-Bench harnesses LLM/LVLM automation to synthesize scene variants and context-aware queries on the fly, which both minimizes over-fitting and drives annotation cost toward zero.  
This high level of automation also makes the suite highly scalable: new lighting regimes or robot behaviours can be added with a few changes to the configuration, so even large, diverse experiments remain low-maintenance and domain-agnostic.


\section{Methodology}

It is crucial to have standardized datasets for testing and evaluating different models and algorithms. However, due to the diversity of real-world scenes and scenarios, creating a comprehensive dataset that covers all possible variations and challenges is a complex task. To address this challenge, we designed an automated LLM/LVLM benchmarking pipeline (see Fig.~\ref{fig:full_pipeline}). Our proposed evaluation methodology encompasses multiple facets to thoroughly assess the performance of open semantic mapping methods.

One of the key ideas of this work is to analyze the sensitivity of SOTA approaches to condition modifications and to identify which methods are most significantly impacted by these changes in terms of semantic segmentation and knowledge graph quality.

In particular, we introduce an approach for synthetically reproducing lighting conditions in a simulation environment, such as placing the primary light source on the camera, introducing dynamic lighting that varies along the robot's path, and removing external light sources. Tweaking these parameters allows for the creation of a diverse set of customized scenes tailored to specific testing scenarios. For example, different times of day can be simulated by adjusting the lighting settings.

Additionally, we vary robotic agent reference velocities and automatically generate movement trajectories within the scene, with random initialization of start and goal poses. This allows for the simulation of realistic agent behaviors and interactions with the environment and provides a more dynamic and challenging testing environment.

Each semantic map is compared against ground truth data, with the baseline configuration serving as the reference performance point. Deviations from this baseline highlight the sensitivity of methods to environmental changes.


\subsection{Sematic Segmentation Evaluation}

\begin{figure*}[htb]
    \centering
    \includegraphics[width=\textwidth]{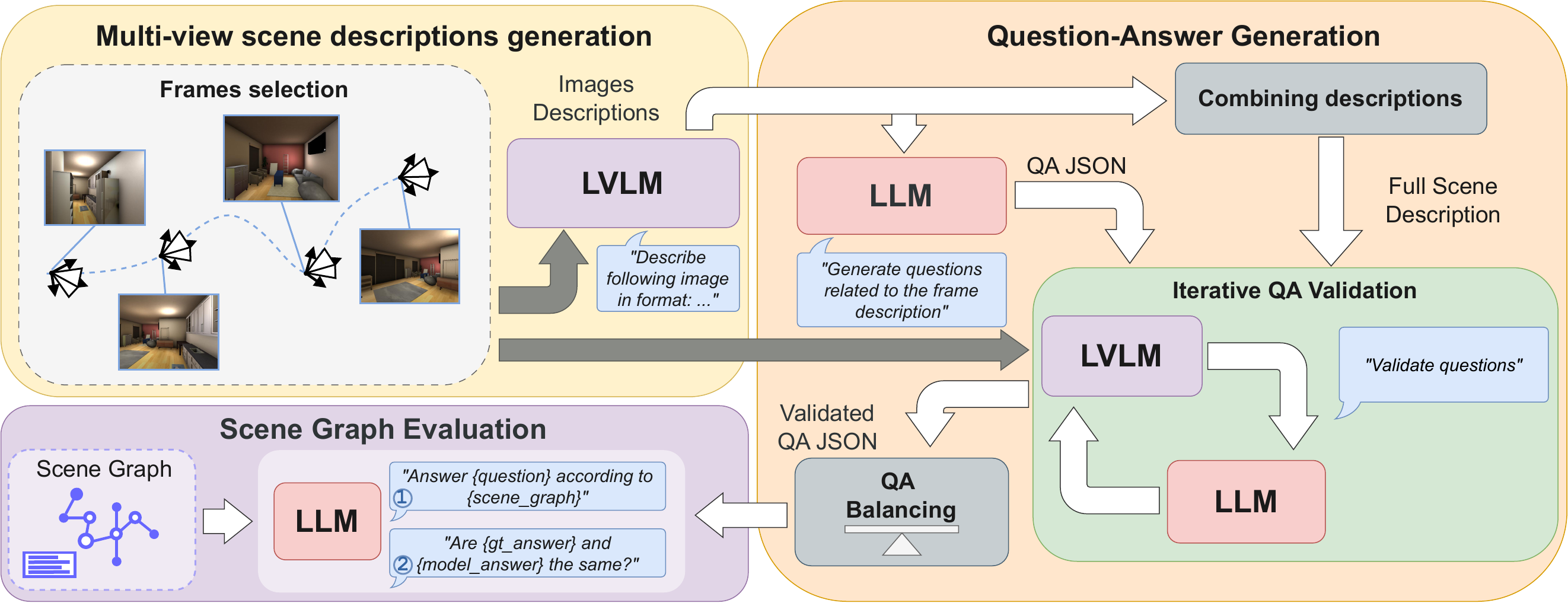}
    \caption{\textit{VQA Pipeline Diagram}. We utilize LLM and LVLM to get scene frames descriptions and construct a set of questions and corresponding ground truth answers, which are additionally validated and balanced in order to avoid usage of ambiguous questions for further evaluation of the scene graph.}
    \label{fig:vqa_pipeline}
\end{figure*}

We evaluate segmentation quality by assessing how well methods segment 3D scenes using our datasets generated using Habitat AI~\cite{szot2021habitat} simulator. Ground truth semantic point clouds are reconstructed using depth images, semantic images, and camera poses. Following the evaluation pipeline from ConceptGraphs~\cite{gu2024conceptgraphs}, we establish a unique closed vocabulary for each scene to generate predictions and employ nearest neighbor matching between predicted and ground truth point clouds. The \textit{mAcc} and \textit{f-mIoU} metrics were chosen to evaluate the segmentation quality.


\subsection{Semantic Graph Evaluation}

To assess the consistency and accuracy of semantic graphs, we developed the question-answer generation pipeline, presented in Figure~\ref{fig:vqa_pipeline}.

We sample frames along the movement within a previously generated test scene. For each key frame, we employ LVLM to generate scene descriptions and further construct a set of questions, each targeting specific aspects of scene understanding:
\begin{enumerate}
    \item Binary General – Yes/No questions about the presence of objects and general scene characteristics (e.g., ``\texttt{Is there a blue sofa?}'');
    \item Binary Existence-Based – Yes/No questions designed to track false positives by querying non-existent objects (e.g., ``\texttt{Is there a piano?}'');
    \item Binary Logical – Yes/No questions with logical operators such as AND/OR (e.g., ``\texttt{Is there a chair AND a table?}'');
    \item Measurement – questions requiring numerical answers related to object counts or scene attributes (e.g., ``\texttt{How many windows are present?}'');
    \item Object Attributes – queries about object properties, including color, shape, and material (e.g., ``\texttt{What color is the door?}'');
    \item Object Relations (Functional) – questions about functional relationships between objects (e.g., ``\texttt{Which object supports the table?}'');
    \item Object Relations (Spatial) – queries about spatial placement of objects within the scene (e.g., ``\texttt{What is in front of the staircase?}'');
    \item Comparison – questions that compare object properties such as size, color, and position (e.g., ``\texttt{Which is taller: the bookshelf or the lamp?}'').
\end{enumerate}

Scene captions from key-frames are merged, an LLM then filters duplicates, fixes ambiguities, and balances object frequencies before evaluation. Duplicate questions are filtered, and answer ambiguities are resolved, ensuring logical consistency across frames. A balancing mechanism prevents object over-representation.

We evaluate answer correctness using the \textit{Answering Accuracy} metric. For binary and measurement-based questions, answers are directly compared to the ground truth. Attribute- and search-related questions' correctness is determined through the LLM-based semantic evaluation. LLM assesses whether the predicted answer aligns with the ground truth, returning a structured JSON output. The final \textit{Accuracy} is computed as the ratio of correct answers to the total number of questions. Prompts enforce structured captions, balanced question types, and reject out-of-scope or ambiguous queries.

The suggested methodology allows adapting test conditions by changing agent reference trajectories, light conditions along the path, and other modifications of a scene, as well as questions from the VQA pipeline.

\section{Experiments}

\subsection{Experiments Organization}

The first step in conducting experiments is data preparation, which involves loading 3D scene models, and configuring lighting, agent, and path. The latter can be done within Habitat Sim~\cite{szot2021habitat} by adjusting the corresponding JSON files.

Two datasets have been deployed for testing in Habitat-Sim environment~\cite{szot2021habitat}. The first one is a synthetic Replica CAD~\cite{szot2021habitat} dataset that was selected because of its flexibility in configuring test conditions and the availability of precise ground truth for both object positions and scene semantics. The second one is a photo-realistic Habitat-Matterport 3D Research Dataset (HM3D)~\cite{ramakrishnan2021habitat}, which is one of the largest collections of high-resolution indoor 3D scans.

We assigned sets of 22 scenes from the ReplicaCAD dataset and 8 scenes from the HM3D dataset, each with distinct test condition configurations:
\begin{itemize}
    \item Pairs of start and goal robot poses were randomly sampled from the available navigation mesh to ensure broad coverage of each scene;
    \item Collision-free path between terminal states has been defined relying on a GreedyGeodesicFollower motion planner available in Habitat-Sim;
    \item Lighting conditions have been set to the following scenarios:
    \begin{itemize}
        \item \textit{"baseline"} configuration uses the static, not-uniformly distributed light sources available as a default scenario for the Replica CAD dataset only with baseline;
        \item \textit{"nominal lights"} configuration relies on the mesh itself emitting light without any added light sources. This means using baked lighting for HM3D and uniformly glowing meshes for ReplicaCAD;
        \item \textit{"camera light"} introduced extra directed light source attached to the camera; 
        \item \textit{"dynamic lighting"} corresponds to changing light conditions along robot path;
    \end{itemize}
    \item Nominal value of robot's constant reference velocities along path were set to 0.75 m/s for transition and 0.8 rad/s for rotation, while for a \textit{"velocity"} modification, we doubled these values for ReplicaCAD \textit{"baseline"} and HM3D \textit{"nominal lights"} conditions;
\end{itemize}

To implement LLM/LVLM blocks within the VQA pipeline (Fig.~\ref{fig:vqa_pipeline}), we used Google Gemini 2.0-Flash~\cite{team2024gemini}.

The VQA pipeline generated on average \textit{184} and \textit{85} questions for each scene of ReplicaCAD and HM3D, which gives us in total \textit{4055} and \textit{681} questions for ReplicaCAD and HM3D, respectively. The average ratio between different questions' categories is the following: \textit{18.8\%} for Binary General, \textit{16.4\%} for Binary Existence-Based, \textit{18.1\%} for Binary Logical, \textit{5.7\%} for Measurement, \textit{16.9\%} for Object Attributes, \textit{1.1\%} for Object Relations - Functional, \textit{18.3\%} for Object Relations - Spatial, \textit{4.8\%} for Comparison.

Functional Relationships were challenging for LLM to interpret correctly, often leading to inconsistent or ambiguous answers. As a result, many of these questions were removed during the validation process and, as a result, we excluded them from the evaluation. 


\subsection{Experimental Results}

\begin{table*}[t]
    \centering
    \small
    \setlength{\tabcolsep}{5pt}
    \caption{Performance on ReplicaCAD}
    \label{tab:replica_cad_combined}
    \begin{tabular}{@{}l|c|cc|cc|cc|cc|cc@{}}
        \toprule
        \multirow{2}{*}{\textbf{Approach}} & \multirow{2}{*}{\shortstack{\textbf{FPS}\\\textbf{(Hz)}}} & 
        \multicolumn{2}{c|}{\textbf{Baseline}} & 
        \multicolumn{2}{c|}{\textbf{Camera Light}} & 
        \multicolumn{2}{c|}{\textbf{Dynamic Lights}} & 
        \multicolumn{2}{c|}{\textbf{Nominal Lights}} & 
        \multicolumn{2}{c}{\textbf{Velocity}} \\
        \cmidrule(lr){3-4} \cmidrule(lr){5-6} \cmidrule(lr){7-8} \cmidrule(lr){9-10} \cmidrule(l){11-12}
        & & \textbf{mAcc} & \textbf{f-mIoU} & \textbf{mAcc} & \textbf{f-mIoU} & \textbf{mAcc} & \textbf{f-mIoU} & \textbf{mAcc} & \textbf{f-mIoU} & \textbf{mAcc} & \textbf{f-mIoU} \\
        \midrule
        ConceptGraphs~\cite{gu2024conceptgraphs} & $0.19^{*}$ 
        & \textbf{29.53} & 28.00 
        & \textbf{29.76} & 22.71 
        & \textbf{24.73} & 14.29 
        & \textbf{26.80} & 21.34 
        & \textbf{28.32} & 27.33 \\
        BBQ~\cite{linok2024beyond}              & $0.93^{*}$ 
        & \underline{25.25} & \underline{36.42} 
        & \underline{26.81} & \underline{29.59} 
        & \underline{24.60} & \underline{32.25} 
        & \underline{25.07} & \underline{33.33} 
        & \underline{26.63} & \underline{32.30} \\
        OpenScene~\cite{peng2023openscene}      & 0.13 
        & 24.17 & \textbf{51.17} 
        & 22.82 & \textbf{48.83} 
        & 22.02 & \textbf{49.05} 
        & 22.16 & \textbf{47.09} 
        & 25.01 & \textbf{55.81} \\
        \bottomrule
    \end{tabular}
    \begin{flushright}
        {\footnotesize $^{*}$ - including scene graph generation}
    \end{flushright}
\end{table*}

\begin{table*}[t]
    \centering
    \caption{Performance on HM3D}
    \small
    \setlength{\tabcolsep}{10pt}
    \label{tab:hm3d}
    \begin{tabular}{@{}l|c|cc|cc|cc@{}}
        \toprule
        \multirow{2}{*}{\textbf{Approach}} & \multirow{2}{*}{\shortstack{\textbf{FPS}\\\textbf{(Hz)}}} & 
        \multicolumn{2}{c|}{\textbf{Nominal Lights}} & 
        \multicolumn{2}{c|}{\textbf{Camera Light}} & 
        \multicolumn{2}{c}{\textbf{Velocity}} \\
        \cmidrule(lr){3-4} \cmidrule(lr){5-6} \cmidrule(l){7-8}
        & & \textbf{mAcc} & \textbf{f-mIoU} & \textbf{mAcc} & \textbf{f-mIoU} & \textbf{mAcc} & \textbf{f-mIoU} \\
        \midrule
        ConceptGraphs~\cite{gu2024conceptgraphs} & $0.16^{*}$ & \underline{20.39} & 17.72 & 16.73 & 15.48 & \underline{19.42} & 17.49 \\
        BBQ~\cite{linok2024beyond}              & $0.92^{*}$ & 18.14 & \underline{23.96} & \underline{17.24} & \underline{24.03} & 18.28 & \underline{24.71} \\
        OpenScene~\cite{peng2023openscene}      & 0.25       & \textbf{20.58} & \textbf{34.19} & \textbf{17.29} & \textbf{29.97} & \textbf{19.77} & \textbf{34.02} \\
        \bottomrule
    \end{tabular}
    \begin{flushright}
        {\footnotesize $^{*}$ - including scene graph generation}
    \end{flushright}
\end{table*}

\begin{figure}[t!]
    \centering
    \includegraphics[width=\linewidth]{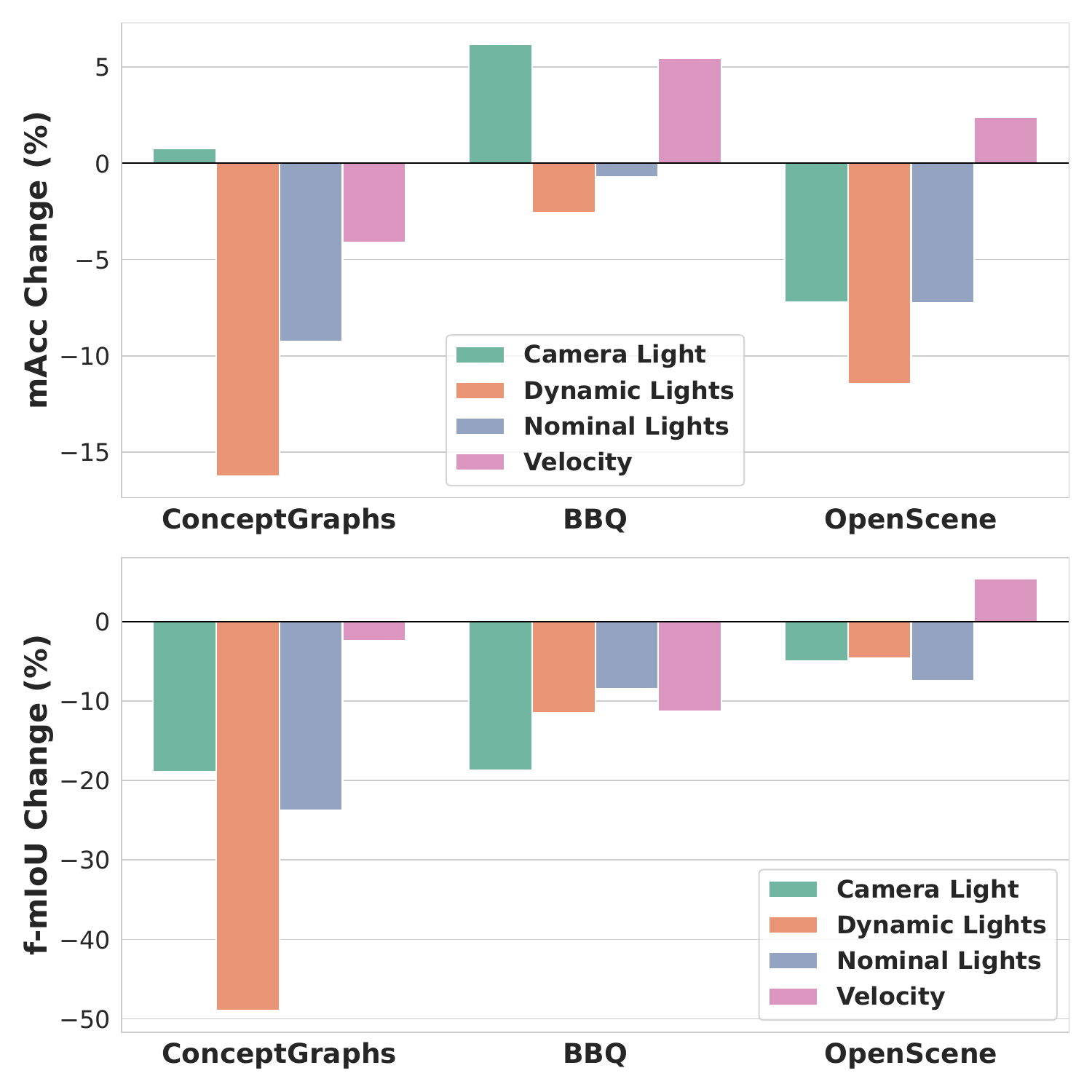}
    \caption{Comparison of changes (\%) in \textit{mAcc} and \textit{f-mIoU} across conditions relative to the baseline on the ReplicaCAD dataset.}
    \label{fig:change_comparison}
\end{figure}

\begin{figure*}[t!]
    \centering    \includegraphics[width=0.85\textwidth]{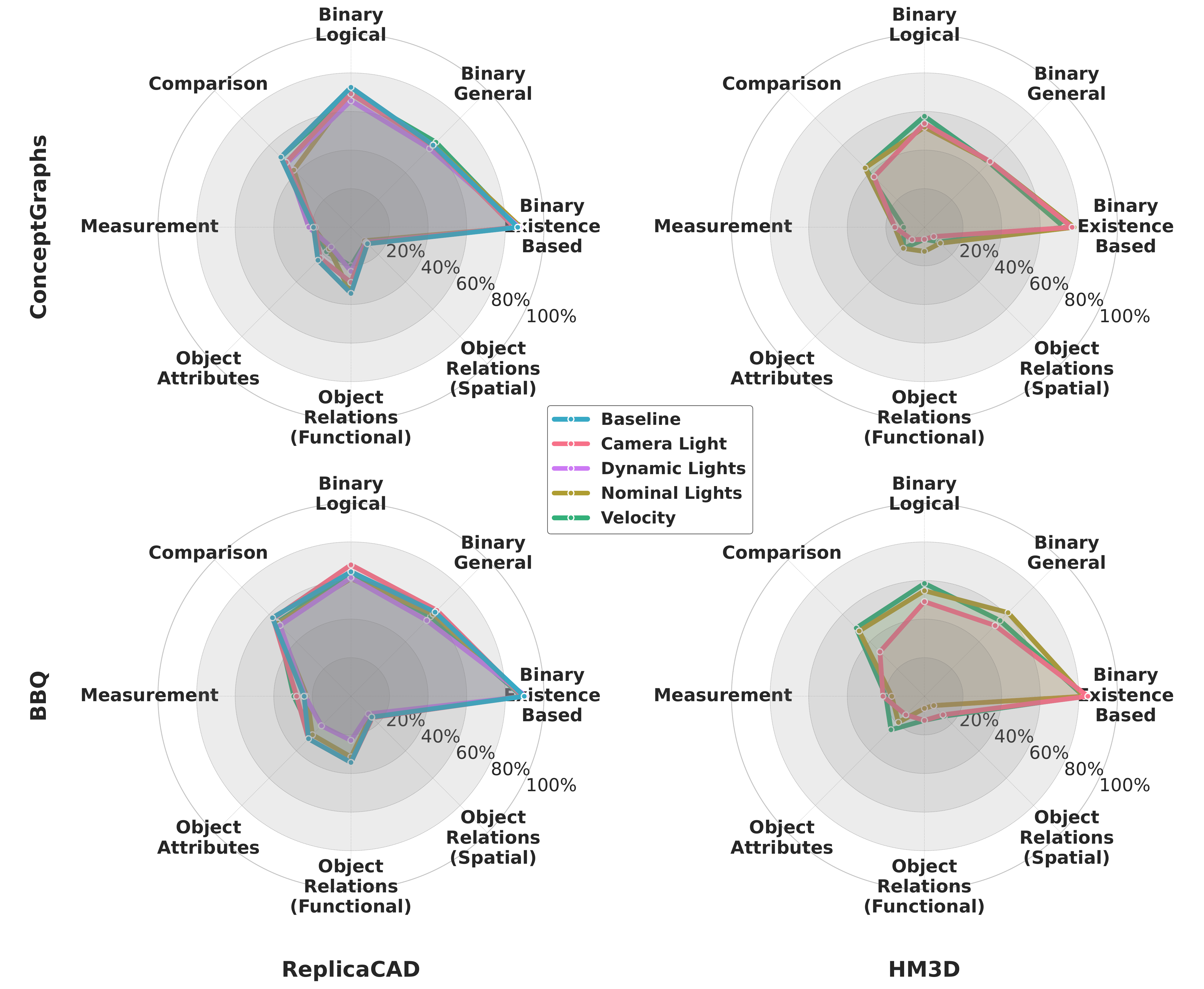} 
    \vspace{-3mm}
    \caption{Comparison of \textit{Answering Accuracy} across ReplicaCAD and HM3D sequences.}
    \label{fig:method_comparison}
\end{figure*}

The evaluation was conducted for three methods: ConceptGraphs~\cite{gu2024conceptgraphs}, OpenScene~\cite{peng2023openscene}, and a very recent BBQ~\cite{linok2024beyond}. We used 2D fused features for OpenScene, as this mode demonstrated the best performance. These methods employ different approaches, so we can perform benchmarking on a wider spectrum.

For more fair comparison between the methods, we replaced LLaVa-7b-v0~\cite{liu2023visual} and GPT4~\cite{achiam2023gpt} originally suggested by ConceptGraphs by LLaVa1.6-7b~\cite{liu2024llavanext} and LLaMa3.1-8B~\cite{dubey2024llama}, also used by BBQ. While LLM and LVLM play an important role in building a semantic graph, it is not a part of reported contributions, and most approaches use it off-the-shelf. Moreover, we wanted to verify performance given a more compact model.

Obtained results for segmentation quality metrics for all considered methods and scenes are presented in Tables~\ref{tab:replica_cad_combined},~\ref{tab:hm3d}, while their relative change under different test conditions is illustrated on a ReplicaCAD dataset by Fig.~\ref{fig:change_comparison}. Final metrics are computed differently for each dataset: for ReplicaCAD, we calculate an overall confusion matrix since it provides a consistent closed vocabulary across all scenes, while for HM3D, we average metrics across individual scenes due to the large number of unique labels and potential label inconsistencies for identical objects across different scenes.

Semantic graph quality assessment has been performed according to the described methodology. Fig~\ref{fig:method_comparison} illustrates VQA accuracy achieved by BBQ and ConceptGraphs methods for different categories of semantic queries. We did not evaluate OpenScene because it does not construct a scene graph.

For BBQ~\cite{linok2024beyond}, questions related to object relations were not considered, as its scene graph construction is query-dependent. Instead, the evaluation focused on assessing the correctness of object presence and attributes, ensuring that the retrieved object set aligns with the given question.

In addition, we measure time consumption using \textit{NVIDIA GeForce RTX 4090} for 3D scene representation in frames per second (\textit{FPS}), which is shown in Tables~\ref{tab:replica_cad_combined},~\ref{tab:hm3d}.


\subsection{Results Analysis}
\label{subsec: results_analysis}



Collecting \textit{mAcc} and \textit{f-mIoU} metrics from testing ConceptGraphs, BBQ, and OpenScene on scenes in ReplicaCAD and HM3D reveals key trade-offs. \textit{f-mIoU} is weighted by class frequency, so voluminous objects like walls and floors matter more, while mAcc gives equal weight to all objects. 

On ReplicaCAD, OpenScene achieves the highest \mbox{\textit{f-mIoU}}, performing best on voluminous objects. ConceptGraphs achieves the highest mAcc, indicating strength in recognizing more rare classes. BBQ performs steadily across both metrics but doesn’t lead in either. Under the velocity setting, OpenScene improves on both metrics, likely due to better filtering with fewer points. Under camera light, both BBQ and ConceptGraphs achieve higher scores in mAcc, likely because the direct lighting emphasizes objects, boosting detection. Also, ConceptGraphs is most sensitive to light conditions.
On HM3D, performance drops across the board, but OpenScene shows the best performance.

Figure~\ref{fig:method_comparison} presents the answering accuracy across different question categories and scene modifications. Binary existence-based questions exhibit high accuracy, indicating that models do not hallucinate non-existent objects. Overall, when binary questions achieve accuracy above 50–60\%, it confirms non-random response patterns.

\begin{figure}[htbp]
    \centering
    \includegraphics[width=\columnwidth]{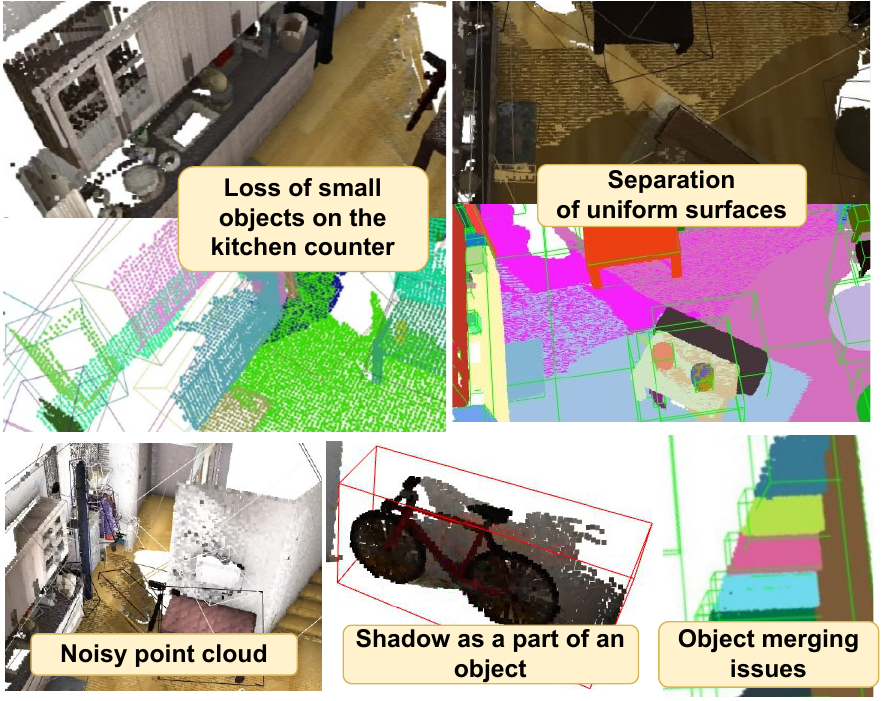}
    \caption{\textit{Illustration of common corner cases in 3D semantic mapping under varying environmental conditions.} Issues include segmentation errors due to illumination changes, loss of small objects during reconstruction, noisy point clouds, misclassification of shadows as objects, and incorrect object merging leading to inflated instance counts in scene graphs.}
    \label{fig:corner_cases}
\end{figure}

The highest accuracy is observed under nominal conditions — the baseline for ReplicaCAD and nominal lighting for HM3D. Variations in lighting and velocity reduce accuracy, especially in object attributes and relationships questions.
Attribute-based questions often fail due to missing material and property details in scene graph captions, leading to incomplete LLM responses.

Taking into account the analysis of metrics, it was also possible to detect corner cases (Fig.~\ref{fig:corner_cases}) under variable lighting conditions, which can lead to a decrease in accuracy.
When using camera light, shadows of objects can be counted as part of the object. Dynamic lighting causes incorrect segmentation of uniform surfaces, such as floors, due to illumination differences. Also, when using only the camera light source, a noisy point cloud was obtained. The tested methods show sensitivity to objects with a similar description (for example, ``\texttt{wall cabinet}'' and ``\texttt{base cabinet}''). Also, when assessing the quality of semantic segmentation, it turned out that the methods show worse results if we use a `\texttt{\_}' instead of a space in the class name.

There is also an object merging issue that affects the answers to Measurement-based questions. For example, in ReplicaCAD, both methods mark each step of the stairs as a separate staircase, e.g.:
{
    \small\texttt{\textbf{Question:} How many staircases are present?} \\
    \texttt{\textbf{Answer:} 1} \\
    \texttt{\textbf{Category:} Measurement} \\
    \texttt{\textbf{Scene Graph Answer:} 15} \\
}


\section{Conclusion}

In this paper, we target a systematic evaluation of open semantic mapping algorithms, in particular their robustness in dynamic environments, e.g., under varying indoor lighting conditions. To accomplish the task, we introduced a dynamically configurable and highly automated LLM/LVLM-powered benchmarking methodology.

Our benchmarking procedure enabled a comprehensive analysis of leading methods, such as OpenScene~\cite{peng2023openscene}, BBQ~\cite{linok2024beyond}, and Concept Graph~\cite{gu2024conceptgraphs}, highlighting the impact of lighting variations on semantic segmentation accuracy and constructed knowledge graphs quality. 

Analysis of the obtained results and corner cases points directions for further research in developing more resilient algorithms capable of maintaining high performance across diverse environments. 

Our solution offers a flexible framework that allows easy extension to evaluate additional capabilities of OSM approaches. OSMa-Bench's adaptability allows it to generate scenes with dynamic objects, evaluate visual grounding, and assess spatial reasoning by expanding the question categories within the existing pipeline. We hope that our code implementation will be useful for further research.



\printbibliography





\appendix

This supplementary material provides additional details and qualitative results that complement the main paper. It clarifies key implementation choices underlying OSMa-Bench and presents representative examples of automatically generated VQA questions and sampled frames used for scene description. It also reports extended statistics on question categories, including their distribution and average counts per scene across datasets, and provides examples of LLM and LVLM prompts employed at different stages of the pipeline. Together, these materials offer deeper insight into the evaluation methodology and support the experimental findings reported in the main manuscript.

\subsection{Removed Questions in VQA Validation}
\label{sec:s_removed_questions}

\begin{figure}[h]
    \centering
    \includegraphics[width=\columnwidth]{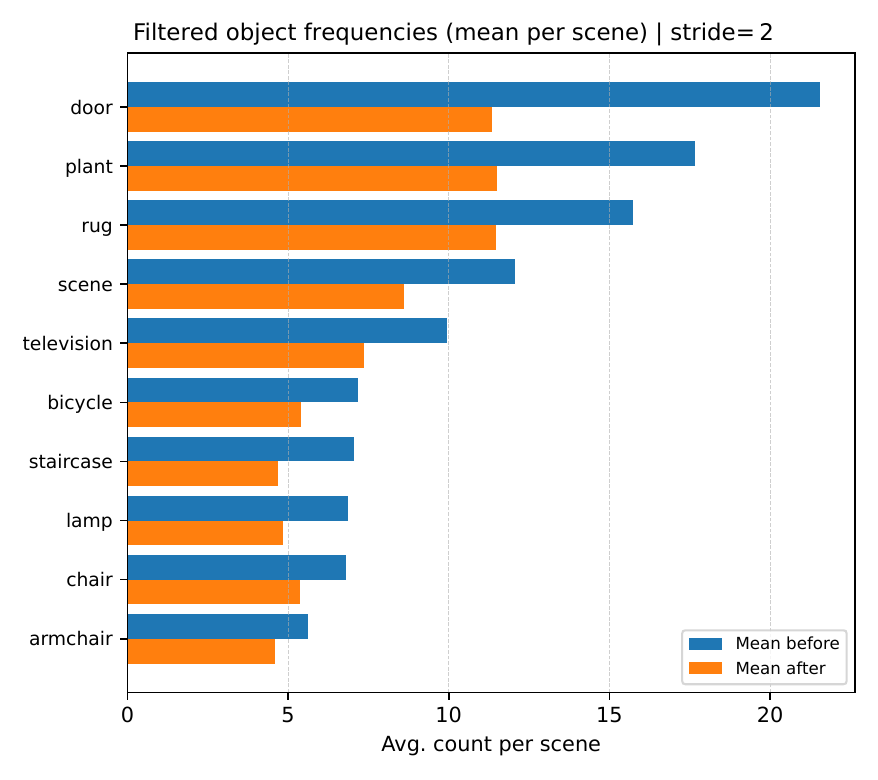}
    \caption{Object frequency distribution before and after validation.}
    \label{fig:objfreq_before_after}
\end{figure}

During the VQA validation stage, a subset of automatically generated question--answer pairs was removed to prevent bias and ambiguity in the evaluation of scene graphs. A common removal case corresponds to exact duplicates and near-duplicates driven by overrepresented objects. When frequently occurring entities dominate the question pool, the generator repeatedly produces semantically redundant queries that do not add new evidence about the map quality. Two representative removed examples (triggered by overused object frequency) are shown below:

\begin{lstlisting}
Question: Is there a television AND a houseplant?
Reason: ('television', 'Overused object frequency')

Question: Is the door white?
Reason: ('door', 'Overused object frequency')
\end{lstlisting}

The impact of frequency-based filtering is illustrated in Fig.~\ref{fig:objfreq_before_after}, which visualizes the object frequency distribution before and after validation. Prior to filtering, a small set of common objects dominates the question set; after filtering, the distribution becomes noticeably flatter, increasing the contribution of less frequent entities and improving diversity, which is essential for open-vocabulary evaluation.

In addition to frequency-based filtering, questions that rely on view-dependent or ill-defined spatial semantics were removed. For example, comparison questions such as the one below implicitly assume a fixed viewpoint, which is incompatible with evaluation over a scene graph representation that is not tied to the camera perspective:

\begin{lstlisting}
{"question": "Which is positioned closer: sofa or cabinet?",
 "answer": "Sofa",
 "category": "Comparison"}
\end{lstlisting}

Similarly, queries involving vague notions such as \textit{background} were excluded because they cannot be consistently interpreted without an explicit observer model: \\

\begin{lstlisting}
{"question": "What is in the background?",
 "answer": "Brown wooden cabinet",
 "category": "Object Relations - Spatial"}
\end{lstlisting}

Representative examples of valid questions per category are provided in the main paper; this section focuses on documenting the validation logic and illustrating how biased, repetitive, or viewpoint-dependent queries are identified and removed, yielding a cleaner and more reliable VQA benchmark.

\subsection{Frame Sampling Along Agent Trajectories}
\label{sec:s_frame_sampling}

To generate diverse and representative question--answer pairs, frames are sampled along the agent trajectory at fixed intervals. This strategy ensures coverage of different viewpoints, object configurations, and spatial relations encountered during navigation, while avoiding excessive redundancy between consecutive frames. The selected frames serve as inputs to the LVLM for scene description, which is later used for VQA generation.

\begin{figure*}[t]
    \centering
    \begin{minipage}{0.24\textwidth}
        \centering
        \includegraphics[width=\linewidth]{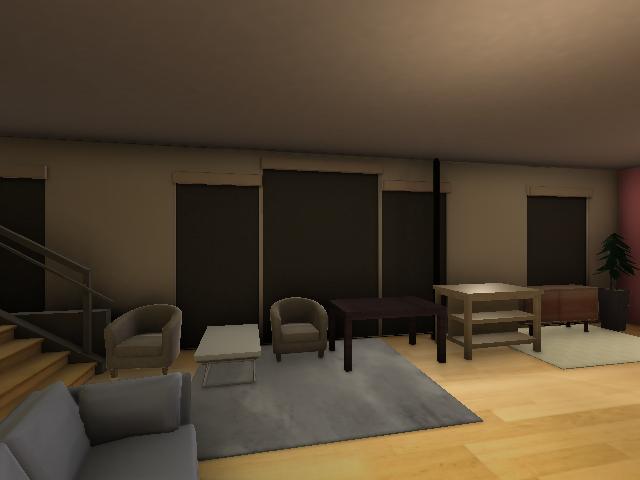}
    \end{minipage}
    \hfill
    \begin{minipage}{0.24\textwidth}
        \centering
        \includegraphics[width=\linewidth]{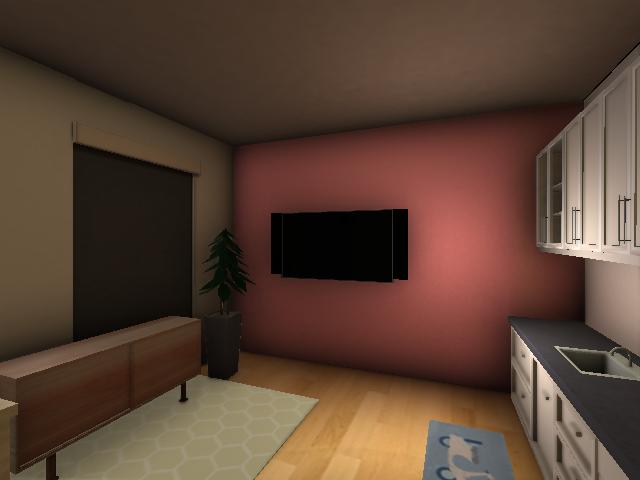}
    \end{minipage}
    \hfill
    \begin{minipage}{0.24\textwidth}
        \centering
        \includegraphics[width=\linewidth]{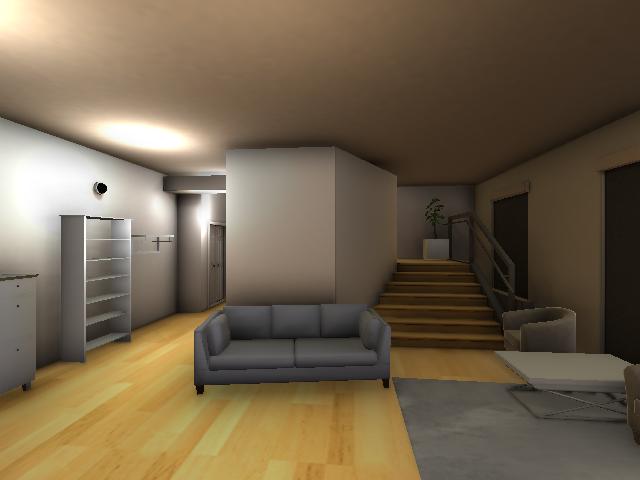}
    \end{minipage}
    \hfill
    \begin{minipage}{0.24\textwidth}
        \centering
        \includegraphics[width=\linewidth]{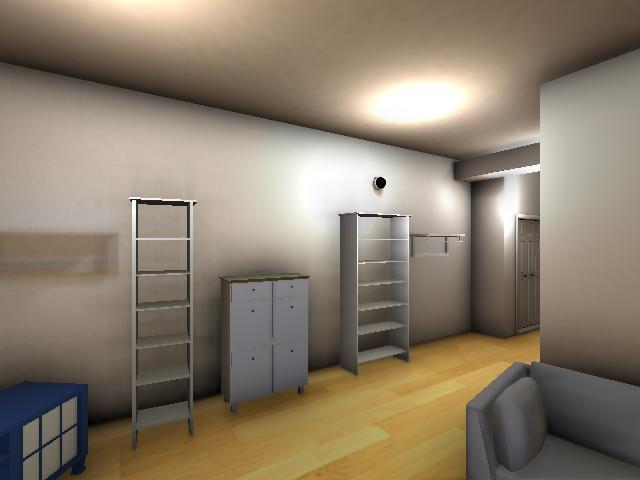}
    \end{minipage}

    \caption{Examples of frames sampled along a single agent trajectory and used for LVLM-based scene description. Frames are selected at different temporal positions to capture complementary viewpoints and object configurations for subsequent VQA generation.}
    \label{fig:frame_sampling_examples}
\end{figure*}

Figure~\ref{fig:frame_sampling_examples} illustrates four representative frames sampled from a single trajectory. Early frames typically capture global scene layout, while later frames provide closer views of objects and reveal additional details relevant for attribute- and relation-based questions. This temporal sampling allows the VQA pipeline to incorporate complementary visual evidence across the trajectory, improving both semantic coverage and question diversity.

\subsection{LLM and LVLM Prompt Design}
\label{sec:s_prompts}

This section documents the prompt design used in OSMa-Bench for scene description, question generation, validation, and answer evaluation. All prompts are derived from a unified configuration file and are designed to enforce structural consistency, avoid ambiguity, and ensure compatibility with open-vocabulary semantic mapping.

\subsubsection{LVLM-Based Scene Description}

Scene descriptions are generated using an LVLM with a strictly structured prompt. The model is instructed to describe \emph{all visible objects} using explicit attributes and pairwise spatial relations, while avoiding vague or view-dependent references. In addition, affordance tags are explicitly attached to each object to support downstream affordance-aware question generation.

A key fragment of the scene description prompt is shown below:

\begin{lstlisting}
Describe the given image and ALL the objects using structured statements.
For EVERY visible object, append affordance tags in square brackets.
Use the verbs: open, pull, push, sit, place, grasp, store, lie.

Use only these two formats:
- {detailed object description} {object} is {spatial relationship} {other object}.
- {detailed object description} {object} [affordance tag] is present in the scene.
\end{lstlisting}

To prevent ill-defined spatial semantics, the prompt explicitly forbids the use of viewpoint-dependent terms:

\begin{lstlisting}
Do not describe objects using vague location references like
"in the background", "on the left", or "at the center".
Always describe relations between exactly two objects.
\end{lstlisting}

Additional annotations are injected to support manipulation-oriented reasoning, including free space in front of objects and vertical clearance above them.

\subsubsection{LLM-Based Question Generation}

Based on the structured scene description, an LLM generates a diverse set of questions spanning multiple semantic categories. The prompt explicitly specifies the allowed question types, target counts per category, and strict rules governing answer formats.

An excerpt of the question generation specification is given below:

\begin{lstlisting}
"categories": [
  {
    "name": "Existence-Positive",
    "count": "3-5",
    "rules": "Questions about objects that DO exist in the scene.
              Answers must be 'Yes'."
  },
  {
    "name": "Measurement",
    "count": "1-3",
    "rules": "Answer must be an Arabic numeral.
              If ambiguous - SKIP the question."
  },
  {
    "name": "Object Affordance",
    "count": "3-5",
    "rules": "Use the affordance verbs extracted earlier."
  }
]
\end{lstlisting}

This explicit categorization enables controlled coverage of object existence, attributes, relations, affordances, and task-relevant properties, while avoiding uncontrolled hallucinations.

\subsubsection{Question Validation and Filtering}

Generated questions undergo a validation stage that cross-checks them against both the input image and the full scene description. The validation prompt removes questions that rely on implicit viewpoints, contain ambiguities, or admit multiple correct answers.

A representative fragment of the validation prompt is shown below:

\begin{lstlisting}
Remove questions that:
- Contradict the image or scene description.
- Require a specific viewpoint (e.g., "Which is closer?").
- Have ambiguous or multiple correct answers.
Return only the filtered questions in JSON format.
\end{lstlisting}

This stage is critical for ensuring that all retained questions are answerable solely from the scene graph representation, without reliance on camera perspective.

\subsubsection{Scene-Graph-Based Answering and Evaluation}

For evaluation, questions are answered using only the constructed scene graph. The answering prompt enforces strict output formatting and prohibits any reasoning beyond the provided graph:

\begin{lstlisting}
Answer the following questions based ONLY on the provided scene graph.
IF Yes/No EXPECTED, ANSWER STRICTLY 'Yes' OR 'No'.
IF COUNTING, ANSWER STRICTLY AS A NUMBER.
\end{lstlisting}

Finally, predicted answers are compared against ground truth using an LLM-based semantic equivalence check, producing a binary correctness label for each question.

Overall, this prompt design enforces a clear separation between perception, question generation, validation, and evaluation, ensuring that the resulting VQA scores reflect the quality and completeness of the underlying semantic map rather than artifacts of language modeling.

\subsection{Distribution of Question Categories}
\label{sec:s_qtypes}

To ensure a balanced and comprehensive evaluation of scene graph quality, the VQA pipeline generates questions across multiple semantic categories, each targeting a distinct aspect of scene understanding. The question set is designed to jointly assess object existence, attributes, spatial and functional relations, quantitative reasoning, and manipulation-related constraints. Such a categorization allows disentangling different failure modes of semantic mapping methods and analyzing how environmental variations affect specific types of reasoning.

Table~\ref{tab:qtype_distribution} summarizes the distribution of question categories and reports the average number of questions per scene for ReplicaCAD and HM3D. While the overall structure of the question set is consistent across datasets, differences in scene complexity and object diversity lead to variations in absolute counts. In particular, ReplicaCAD scenes yield a larger number of questions due to their denser semantic annotations, whereas HM3D scenes result in a more compact but structurally similar distribution. This design ensures that the evaluation remains comparable across datasets while adapting to their inherent characteristics.

\begin{table}[t]
\centering
\small
\setlength{\tabcolsep}{6pt}
\begin{tabular}{lcc}
\toprule
\textbf{Question Category} & \textbf{ReplicaCAD} & \textbf{HM3D} \\
\midrule
Existence-Positive            & 18.8\% & 17.2\% \\
Existence-Negative            & 16.4\% & 18.1\% \\
Binary Logical                & 18.1\% & 16.9\% \\
Object Attributes             & 16.9\% & 17.8\% \\
Object Relations (Spatial)    & 18.3\% & 17.5\% \\
Object Relations (Functional) & 1.1\%  & 0.0\%  \\
Measurement                   & 5.7\%  & 6.3\%  \\
Comparison                    & 4.8\%  & 6.2\%  \\
\midrule
\textbf{Avg. questions / scene} & \textbf{184} & \textbf{85} \\
\bottomrule
\end{tabular}
\caption{Distribution of VQA question categories and average number of questions per scene for ReplicaCAD and HM3D. Percentages indicate the relative share of each category within the validated question set.}
\label{tab:qtype_distribution}
\end{table}

\subsection{Representative Corner Cases}
\label{sec:s_staircase_corner_case}

During the analysis of open semantic mapping methods, several systematic corner cases were identified for ConceptGraphs, revealing limitations related to instance granularity, mask consistency, and semantic aggregation. These failure modes arise at different stages of the pipeline and propagate to scene graph construction and subsequent graph-based reasoning.

\begin{figure}[h]
    \centering
    \includegraphics[width=0.65\columnwidth]{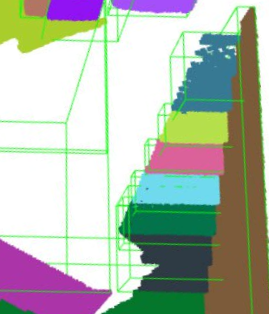}
    \caption{\textit{Staircases} instances created during caption generation.}
    \label{fig:staircase_captions}
\end{figure}

A more severe failure mode concerns semantic over-fragmentation of objects. Figure~\ref{fig:staircase_captions} shows the output, where individual stair steps are detected as separate object instances and each is assigned the same semantic label \textit{staircase}. While this behavior is locally consistent with frame-level visual evidence, it results in multiple semantically identical nodes being introduced into the scene graph for what is physically a single object.

\begin{figure}[h]
    \centering
    \includegraphics[width=\columnwidth]{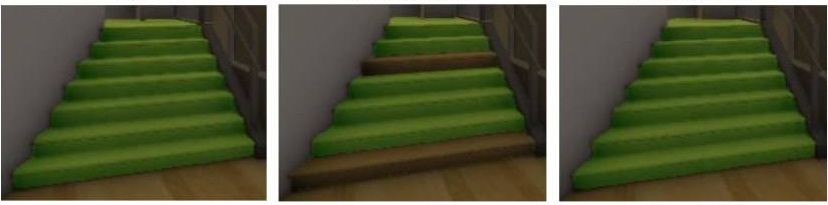}

    \vspace{2mm}
    \includegraphics[width=0.95\columnwidth]{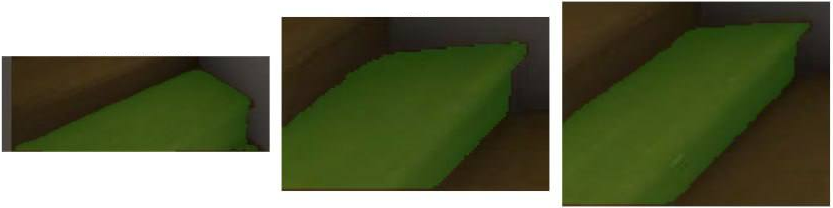}

    \caption{Segmentation masks for staircases at different granularities.
    The top images show a single mask correctly covering the entire staircase, while the bottom images illustrate step-level masks, where each stair step is segmented individually but still associated with the semantic label \textit{staircase}.}
    \label{fig:staircase_masks}
\end{figure}

The corresponding segmentation behavior is shown in Fig.~\ref{fig:staircase_masks}. In the top example, the entire staircase is correctly segmented as a single instance. In contrast, the bottom example presents step-level segmentation masks of the same staircase, where stair step is segmented individually but still associated with the semantic label \textit{staircase}. Although this distinction is not apparent from the semantic label alone, it becomes explicit at the graph level, where each step-level region contributes an independent node with identical semantics.

This mismatch between geometric segmentation granularity and semantic labeling leads to systematic errors during graph-based reasoning. In particular, measurement and counting questions interpret each step-level instance as a separate staircase, resulting in inflated object counts and incorrect answers (Section~\ref{subsec: results_analysis}).

\begin{figure}[h]
    \centering
    \includegraphics[width=0.9\columnwidth]{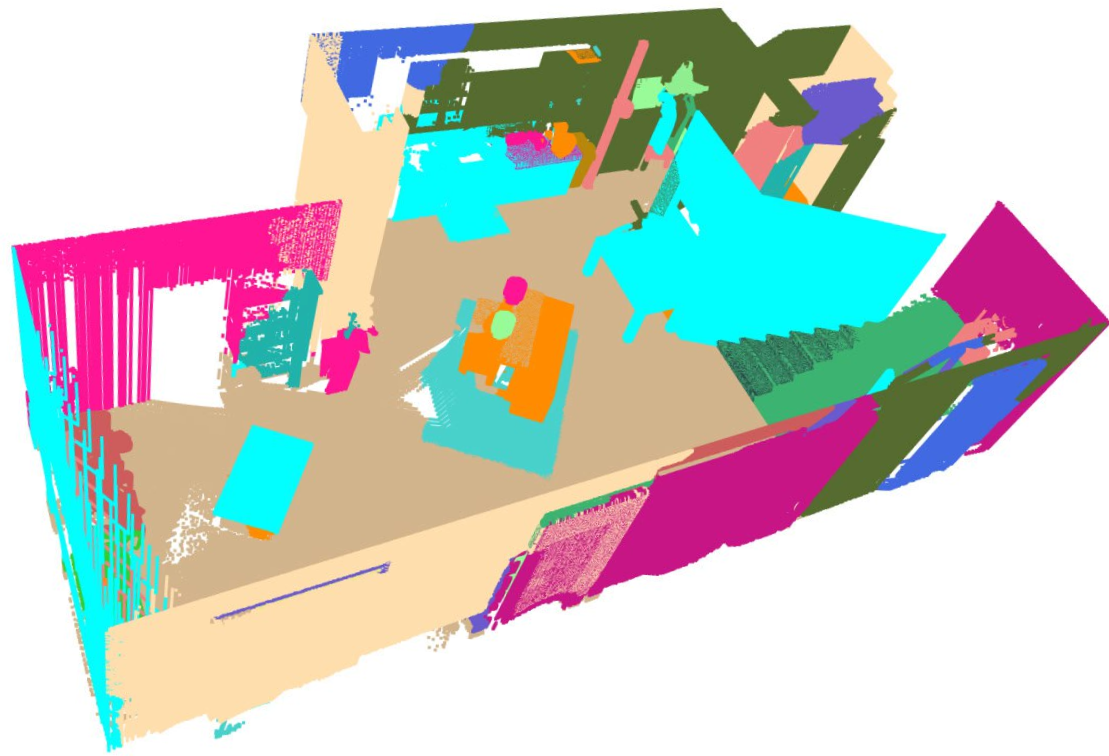}
    \caption{Fragmentation and inconsistent merging of large planar structures in ConceptGraphs}
    \label{fig:wall_ceiling_merging}
\end{figure}

Another recurring issue is observed for large planar structures such as walls and ceilings. As illustrated in Fig.~\ref{fig:wall_ceiling_merging}, these surfaces are often segmented into multiple spatial fragments that are not consistently merged into a single instance. Due to similar appearance and limited geometric cues at fragment boundaries, parts of walls and ceilings may either remain disconnected or be incorrectly grouped with adjacent structures. This leads to unstable graph representations, where a single physical wall or ceiling is represented by several weakly connected nodes.

\begin{figure}[t]
    \centering
    \begin{minipage}[t]{0.29\textwidth}
        \centering
        \includegraphics[height=3.5cm,keepaspectratio]{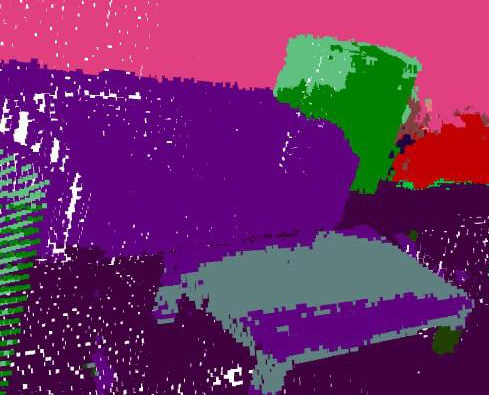}
        \caption*{(a) Sofa mask spreading onto table}
    \end{minipage}
    \begin{minipage}[t]{0.18\textwidth}
        \centering
        \includegraphics[height=3.5cm,keepaspectratio]{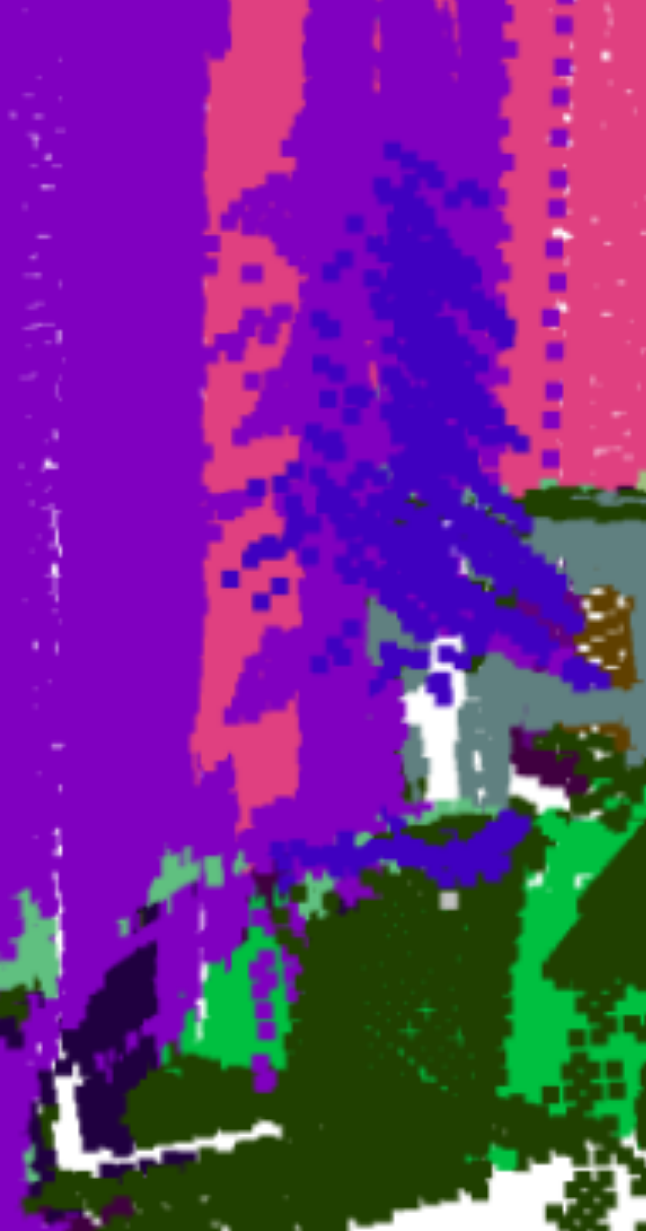}

        \caption*{\centering(b) Wall mask spreading onto plant}
    \end{minipage}

    \caption{Examples of fuzzy mask propagation across adjacent objects. Due to insufficient separation of visual embeddings, object masks extend beyond true boundaries, contaminating neighboring instances.}
    \label{fig:fuzzy_masks}
\end{figure}

One class of errors is related to fuzzy mask propagation, where object masks extend beyond their true geometric boundaries and partially cover neighboring objects. As illustrated in Fig.~\ref{fig:fuzzy_masks}(a)--(b), masks corresponding to large planar structures or visually dominant objects may spread onto adjacent furniture or plants. This effect contaminates object boundaries and leads to incorrect assignment of visual features, particularly under challenging lighting conditions or low-contrast boundaries.

Overall, these corner cases show that in ConceptGraphs, semantic labels alone are insufficient to enforce coherent object-level representations. When geometric segmentation produces fragments that are either too fine-grained or poorly merged, physically continuous structures may appear multiple times in the scene graph, directly affecting downstream reasoning and evaluation.

\subsection{Examples of Lighting Configurations}
\label{sec:s_lighting_examples}

\begin{figure*}[htbp!]
    \centering
    \includegraphics[width=\textwidth]{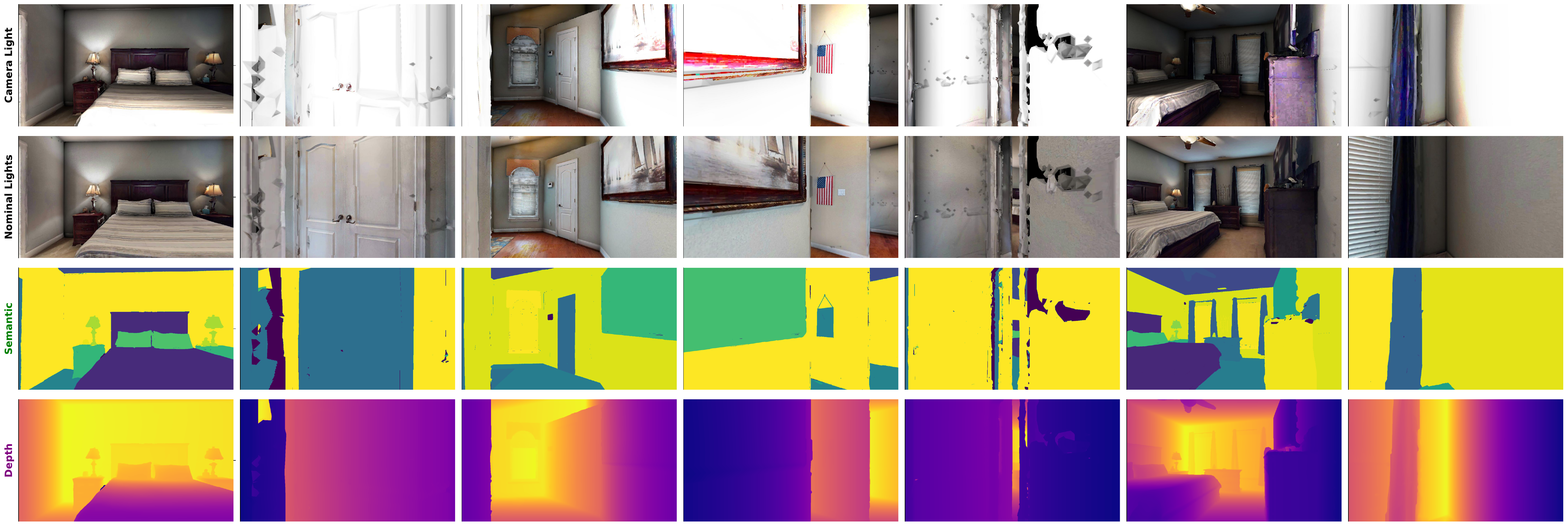}
    \caption{Examples of different lighting configurations used in OSMa-Bench for HM3D dataset: baseline, nominal lights, camera-mounted light, and dynamic lighting along the trajectory.}
    \label{fig:lighting_examples_hm3d}
\end{figure*}

To analyze robustness under realistic visual variability, \textit{OSMa-Bench} evaluates each scene under multiple lighting configurations. These configurations are designed to emulate common indoor scenarios encountered by mobile robots and to isolate the impact of illumination on perception and mapping.

\begin{figure*}[ht]
    \centering
    \includegraphics[width=\textwidth]{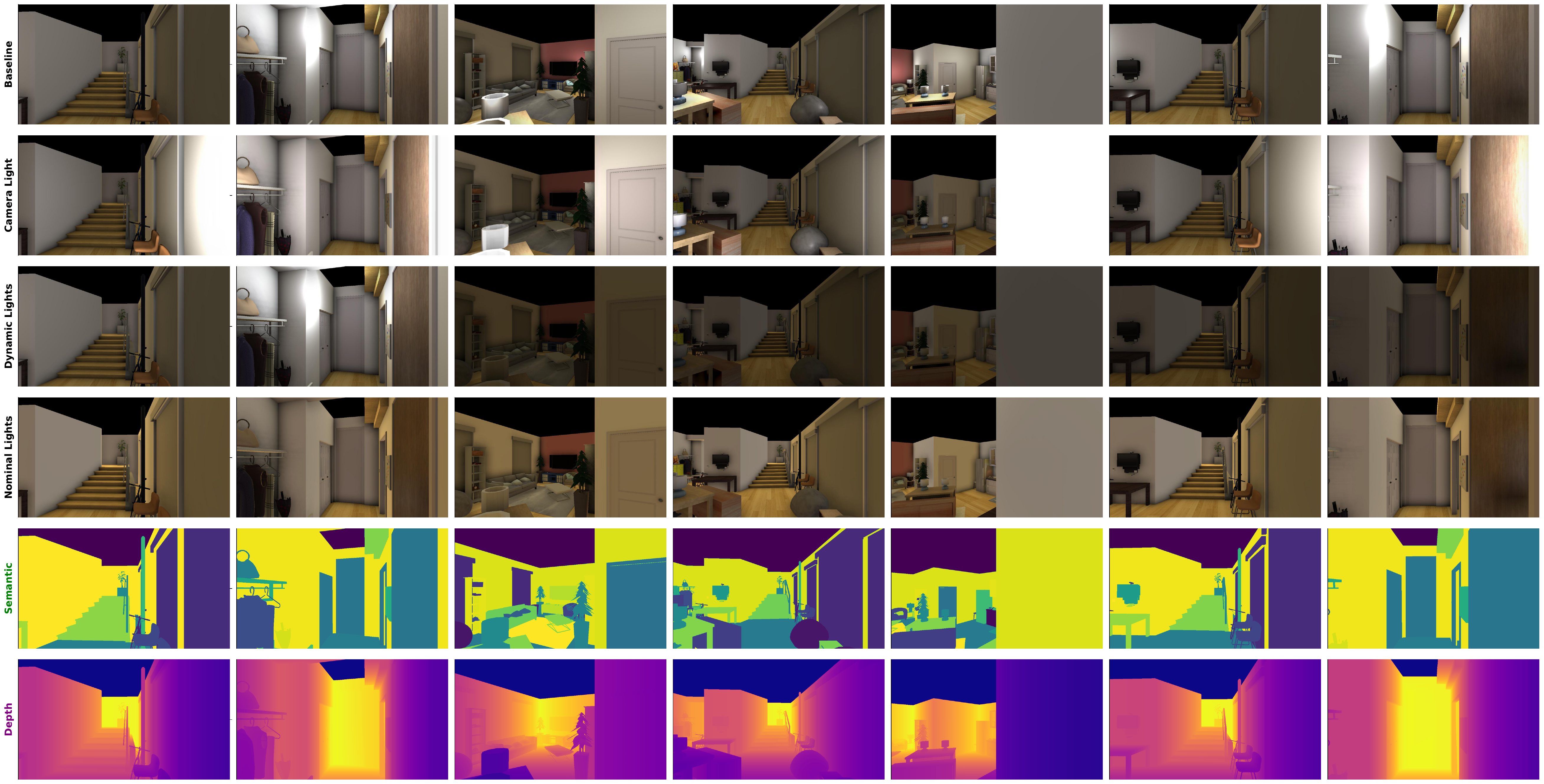}
    \caption{Examples of different lighting configurations used in OSMa-Bench for ReplicaCAD dataset: baseline, nominal lights, camera-mounted light, and dynamic lighting along the trajectory.}
    \label{fig:lighting_examples_replica_cad}
\end{figure*}

Figures~\ref{fig:lighting_examples_replica_cad},~\ref{fig:lighting_examples_hm3d} illustrate representative frames captured under baseline, nominal lights, camera light, and dynamic lighting conditions. While baseline and nominal lighting provide relatively stable visual cues, camera-mounted illumination introduces strong directional shadows and specular highlights. Dynamic lighting further increases scene complexity by altering illumination along the trajectory, often leading to inconsistent segmentation of large planar surfaces such as floors and walls.

These examples demonstrate that changes in lighting alone, without modifying geometry or object layout, can significantly affect both low-level segmentation and high-level semantic reasoning. As a result, lighting variation emerges as a critical factor for evaluating the reliability of open semantic mapping systems.

\end{document}